# Masked Vision and Language Pre-training with Unimodal and Multimodal Contrastive Losses for Medical Visual Question Answering

Pengfei Li[1], Gang Liu[1(✉)], Jinlong He[1], Zixu Zhao[1] and Shenjun Zhong[2(✉)]

[1] College of Computer Science and Technology, Harbin Engineering University, China
`liugang@hrbeu.edu.com`

[2] Monash Biomedical Imaging, Monash University, Australia
`shenjun.zhong@monash.edu`

**Abstract.** Medical visual question answering (VQA) is a challenging task that requires answering clinical questions of a given medical image, by taking consider of both visual and language information. However, due to the small scale of training data for medical VQA, pre-training fine-tuning paradigms have been a commonly used solution to improve model generalization performance. In this paper, we present a novel self-supervised approach that learns unimodal and multimodal feature representations of input images and text using medical image caption datasets, by leveraging both unimodal and multimodal contrastive losses, along with masked language modeling and image text matching as pre-training objectives. The pre-trained model is then transferred to downstream medical VQA tasks. The proposed approach achieves state-of-the-art (SOTA) performance on three publicly available medical VQA datasets with significant accuracy improvements of 2.2%, 14.7%, and 1.7% respectively. Besides, we conduct a comprehensive analysis to validate the effectiveness of different components of the approach and study different pre-training settings. Our codes and models are available at https://github.com/pengfeiliHEU/MUMC.

## 1 Introduction

Medical VQA is a specialized domain of VQA that aims to generate answers to natural language questions about medical images. It is very challenging to train deep learning based medical VQA models from scratch, since the medical VQA datasets available for research are relatively small in scale. Many existing works are proposed to leverage pre-trained visual encoders with external datasets to solve downstream medical VQA tasks, such as utilizing denoising autoencoders [1] and meta-models [2]. These methods mainly transfer feature encoders that are separately pre-trained on unimodal (image or text) tasks.

Unlike unimodal pretraining approaches, both image and text feature presentations can be enhanced by learning through the visual and language interactions, given relatively richer resources of medical image caption datasets [3-5]. Liu et al. followed the work of MOCO [19] that trained teacher model for visual encoder via contrastive loss of different image views (by data augmentations) to improve the generalization of medical VQA [6]. Eslami et al. utilized CLIP [7] for visual model initialization, and learned cross-modality representations from medical image-text pairs by maximizing the cosine similarity between the extracted features of medical images and their corresponding captions [8]. Cong et al. devised an innovative framework, which featured a semantic focusing module to emphasize image regions that were pertinent to the caption and a progressive cross-modality comprehension module that iteratively enhanced the comprehension of the correlation between the image and caption [9]. Chen et al. proposed a medical vision language pre-training approach that used both masked image modelling and masked language modelling to jointly learn representations of medical images and their corresponding descriptions [10]. However, to the best of our knowledge, there have been no existing methods that explore learning both unimodal and multimodal features at the pre-training stage for downstream medical VQA tasks.

In this paper, we proposed a new self-supervised vision language pre-training (VLP) approach that applied **M**asked image and text modeling with **U**nimodal and **M**ultimodal **C**ontrastive losses (MUMC) in the pre-training phase for solving downstream medical VQA tasks. The model was pretrained on image caption datasets for aligning visual and text information, and transferred to downstream VQA datasets. The unimodal and multimodal contrastive losses in our work are applied to (1) align image and text features; (2) learn unimodal image encoders via momentum contrasts of different views of the same image (i.e. different views are generated by different image masks); (3) learn unimodal text encoder via momentum contrasts. We also introduced a new masked image strategy by randomly masking the patches of the image with a probability of 25%, which serves as a data augmentation technique to further enhance the performance of the model. Our approach outperformed existing methods and sets new benchmarks on three medical VQA datasets [11-13], with significant enhancements of 2.2%, 14.7%, and 1.7% respectively. Besides, we conducted an analysis to verify the effectiveness of different components and find the optimal masking probability. We also conducted a qualitative analysis on the attention maps using Grad-CAM [14] to validate whether the corresponding part of the image is attended when answering a question.

## 2 Methods

In this section, we provide the detailed description of the proposed approach, which includes the network architectures, self-supervised pre-training objectives, and the way to fine-tune on downstream medical VQA tasks.

## 2.1 Model Architecture

In the pre-training phase, the network architecture comprises an image encoder, a text encoder, and a multimodal encoder, which are all based on the transformer architecture [15]. As shown in Fig. 1(a), the image encoder leverages a 12-layer Vision Transformer (ViT) [16] to extract visual features from the input images, while the text encoder employs a 6-layer transformer which is initialized by the first 6 layers of pre-trained BERT [17]. The last 6 layers of BERT are utilized as the multimodal encoder and incorporated cross-attention at each layer, which fuses the visual and linguistic features to facilitate learning of multimodal interactions. The model is trained on medical image-caption pairs. An image is partitioned into patches of size $16 \times 16$, and 25% of the patches are randomly masked. The remaining unmasked image patches are converted into a sequence of embeddings by an image encoder. The text, i.e. the image caption is tokenized into a sequence of tokens using a WordPiece [18] tokenizer and fed into the BERT-based text encoder. In addition, the special tokens, [CLS] are appended to the beginning of both the image and text sequence.

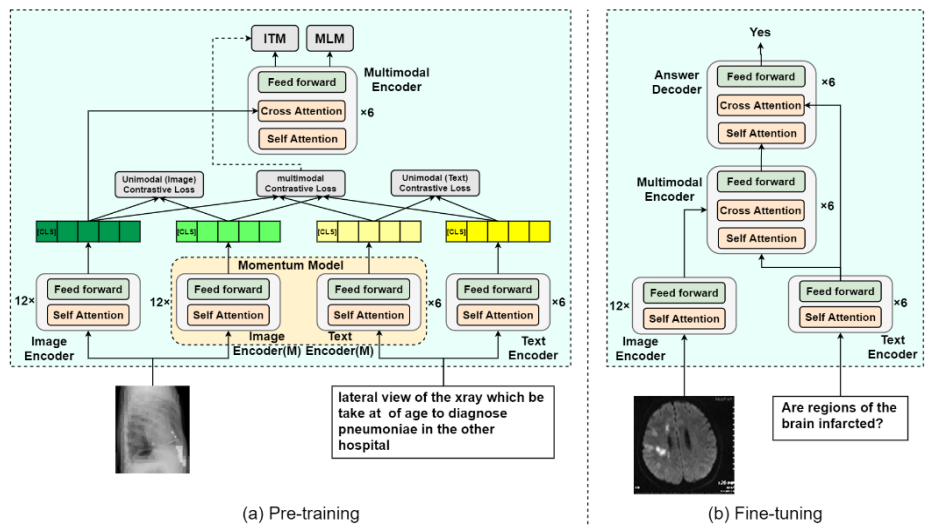

**Fig. 1.** Overview of the network architecture in both pre-training and fine-tuning phases.

To transfer the models trained on image caption datasets to the downstream medical VQA tasks, we utilize the weights from the pre-training stage to initialize the image encoder, text encoder and multimodal encoder, as shown in Fig. 1(b). To generate answers, we add an answering decoder with a 6-layer transformer-based decoder to the model, which receives the multimodal embeddings and output text tokens. A [CLS] token serves as the initial input token for the decoder, and a [SEP] token is appended to signify the end of the generated sequence. The downstream VQA model is fine-tuned via the masked language model (MLM) loss [17], using ground-truth answers as targets.

## 2.2 Unimodal and Multimodal Contrastive Losses

The proposed self-supervised objective attempts to capture the semantic discrepancy between positive and negative samples across both unimodal and multimodal domains at the same time. The unimodal contrastive loss (UCL) aims to differentiate between examples of one modality, such as images or text, in a latent space to make similar examples close. And the multimodal contrastive loss (MCL) learns the alignments between both modalities by maximizing the similarity between images and their corresponding text captions, while separating from the negative examples. In the implementation, we maintain two momentum models for image and text encoders respectively to generate different perspectives or representations of the same input sample, which serve as positive samples for contrastive learning.

In detail, we denote the image and caption embeddings from the unimodal image encoder and text encoder as $v_{cls}$ and $t_{cls}$, which are further processed through the transformations $g_v$ and $g_t$, to normalize and map the image and text embeddings to be lower-dimensional representations. The embeddings are inserted into a lookup table, and only the most recent 65,535 pairs of image-text embedding are stored for contrastive learning. We utilize the momentum update technique originally proposed in Mo-Co [19], which is updated every $k$ iterations where k is a hyperparameter. We denote the ground-truth one-hot similarity by $y_{i2i}(V)$, $y_{t2t}(T)$, $y_{i2t}(V)$, and $y_{t2i}(T)$, where the probability of negative pairs is 0 and the probability of the positive pair is 1. The unimodal contrastive losses and multimodal contrastive losses can be defined as the cross-entropy $H$ given as follows:

$$L_{ucl} = \frac{1}{2}\mathbb{E}_{(V,T)\,D}\left[H\left(y_{i2i}(V),\frac{\exp(s(V,V_i)/\tau)}{\sum_{n=1}^{N}\exp(s(V,V_i)/\tau)}\right) + H\left(y_{t2t}(T),\frac{\exp(s(T,T_i)/\tau)}{\sum_{n=1}^{N}\exp(s(T,T_i)/\tau)}\right)\right] \quad (1)$$

$$L_{mcl} = \frac{1}{2}\mathbb{E}_{(V,T)\,D}\left[H\left(y_{i2t}(V),\frac{\exp(s(V,T_i)/\tau)}{\sum_{n=1}^{N}\exp(s(V,T_i)/\tau)}\right) + H\left(y_{t2i}(T),\frac{\exp(s(T,V_i)/\tau)}{\sum_{n=1}^{N}\exp(s(T,V_i)/\tau)}\right)\right] \quad (2)$$

where $s$ denotes cosine similarity function, $s(V,V_i) = g_v(v_{cls})^T g_v(v_{cls})_i$, $s(T,T_i) = g_t(t_{cls})^T g_t(t_{cls})_i$, $s(V,T_i) = g_v(v_{cls})^T g_t(t_{cls})_i$, $s(T,V_i) = g_t(t_{cls})^T g_v(v_{cls})_i$ and $\tau$ is a learnable temperature parameter.

## 2.3 Image Text Matching

We adopt the image text matching (ITM) strategy similar to prior works [20, 21] as one of the training objectives, by creating a binary classification task with negative text labels randomly sampled from the same minibatch. The joint representation of the image and text are encoded by the multimodal encoder, and utilized as input to the binary classification head. The ITM task is optimized using the cross-entropy loss:

$$\mathcal{L}_{itm} = \mathbb{E}_{(V,T)D}H(y_{itm},p_{itm}(V,T)) \quad (3)$$

the function $H(,)$ represents a cross-entropy computation, where $y_{itm}$ denotes the ground-truth label and $p_{itm}(V,T)$ is a function for predicting the class.

### 2.4 Masked Language Modeling

Masked Language Modeling (MLM) is another pre-trained objective in our approach, that predicts masked tokens in text based on both the visual and unmasked contextual information. For each caption text, 15% of tokens are randomly masked and replaced with the special token, [MASK]. Predictions of the masked tokens are conditioned on both unmasked text and image features. We minimize the cross-entropy loss for MLM:

$$\mathcal{L}_{mlm} = \mathbb{E}_{(V,\hat{T})D} H(y_{mlm}, p_{mlm}(V, \hat{T})) \tag{4}$$

where $H(,)$ is a cross-entropy calculation, $\hat{T}$ denotes the masked text token, $y_{mlm}$ represents the ground-truth of the masked text token and $p_{mlm}(V, \hat{T})$ is the predicted probability of a masked token.

### 2.5 Masked Image Strategy

Besides the training objectives, we introduce a masked image strategy as a data augmentation technique. In our experiment, input images are partitioned into patches which are randomly masked with a probability of 25%, and only the unmasked patches are passed through the network. Unlike the previous methods [10, 22], we do not utilize reconstruction loss [23], but use this only as a data augmentation method. This enables us to process more samples at each step, resulting in a more efficient pre-training of vision-language models with a similar memory footprint.

## 3 Experiments

### 3.1 Datasets

Our model is pre-trained on three datasets: ROCO [3], MedICaT [4], and the ImageCLEF2022 Image Caption Dataset [5]. ROCO comprises over 80,000 image-caption pairs. MedICaT includes over 217,000 medical images and their corresponding captions. ImageCLEF2022 is another well-known dataset that has nearly 90,000 pairs of medical images and captions.

For the downstream medical VQA task, we fine-tune and validate the model on three public medical VQA datasets: VQA-RAD [11], PathVQA [12] and SLAKE [13]. VQA-RAD has 315 radiology images with 3064 question-answer pairs, with 451 pairs used for testing. SLAKE has 14,028 pairs of samples which are further divided into 70% training, 15% validation, and 15% testing subsets. PathVQA is the largest dataset, containing 32,799 pairs of data that are split into training (50%), validation (30%), and test (20%) sets.

There are two types of questions: closed-ended questions that have limited answer choices (e.g. "yes" or "no") and open-ended questions that VQA models are required to generate answers in free text, which are more challenging.

### 3.2 Implementation Details

Our method was implemented in Python 3.8 and PyTorch 1.10. The experiments were conducted on a server with an Intel Xeon(R) Platinum 8255C and 2 NVIDIA Tesla V100 GPUs with 32GB memory each. We pre-trained our model on three medical image caption datasets for 40 epochs with a batch size of 64. AdamW [24] optimizer was used with a weight decay of 0.002 and an initial learning rate of $1e^{-4}$, which decayed to $2e^{-5}$ by following the cosine schedule. We utilized randomly cropped images of $256 \times 256$ resolution as input, and also applied RandAugment to augment more training samples [25].

For downstream medical VQA tasks, we fine-tuned our model for 30 epochs with a batch size of 8. We used the AdamW optimizer with a reduced learning rate of $2e^{-5}$, which decayed to $1e^{-8}$. Besides, we increased image inputs from a resolution of $256 \times 256$ to $384 \times 384$ and interpolated the positional encoding following [16].

### 3.3 Comparison with the State-of-the-arts

We performed a comparative evaluation of our model against the existing approaches [10, 26] on three benchmark datasets, VQA-RAD, PathVQA and SLAKE. Consistent with previous research [1, 2, 6, 8-10, 26, 27], we adopt accuracy as the performance metric. We treated VQA as a generative task by calculating similarities between the generated answers and candidate list answers, selecting the highest score as the final answer. As shown in Table 1, our approach outperformed all other methods on all the three datasets in terms of overall performance, and yielded the best accuracy for open-ended or closed-ended answers. On the VQA-RAD dataset [11], our method achieved an absolute margin of 2.2% overall over the current state-of-the-art method, M3AE, with improvements of 4.3% and 0.7% on open-ended and closed-ended answers respectively.

**Table 1.** Comparisons with the state-of-the-art methods on the VQA-RAD, PathVQA and SLAKE test set.

| Methods | VQA-RAD | | | PathVQA | | | SLAKE |
|---|---|---|---|---|---|---|---|
| | Open | Closed | Overall | Open | Closed | Overall | Overall |
| MEVF [1] | 43.9 | 75.1 | 62.6 | 8.1 | 81.4 | 44.8 | 78.6 |
| MMQ [2] | 52.0 | 72.4 | 64.3 | 11.8 | 82.1 | 47.1 | - |
| VQAMix [27] | 56.6 | 79.6 | 70.4 | 13.4 | 83.5 | 48.6 | - |
| AMAM [26] | 63.8 | 80.3 | 73.3 | 18.2 | 84.4 | 50.4 | - |
| CPRD [6] | 61.1 | 80.4 | 72.7 | - | - | - | 82.1 |
| PubMedCLIP [8] | 60.1 | 80.0 | 72.1 | - | - | - | 80.1 |
| MTL [9] | 69.8 | 79.8 | 75.8 | - | - | - | 82.5 |
| M3AE [10] | 67.2 | 83.5 | 77.0 | - | - | - | 83.2 |
| **MUMC (Ours)** | **71.5** | **84.2** | **79.2** | **39.0** | **90.4** | **65.1** | **84.9** |

On the largest dataset, PathVQA [12], our method significantly outperformed the previous state-of-the-art model, AMAM [26], by a substantial margin with improvements of 20.8%, 6.0% and 14.7% on the closed-ended, open-ended, and overall answers, respectively. Moreover, on the SLAKE dataset [13], the proposed approach exhibited superior performance compared to the existing state-of-the-art model, M3AE, by a margin of 1.7% in terms of overall answer accuracy.

### 3.4 Ablation Study

To further verify the effectiveness of the proposed methods in learning multimodal representations, we conducted an ablation study across all three medical VQA datasets. Table 2 shows the overall performance of the medical VQA tasks using various pre-training approaches. Compared to the baseline pre-training tasks (i.e., MLM + ITM), integrating either UCL or MCL significantly improved the performance of the pre-trained model across all medical VQA datasets. Notably, the simultaneous use of UCL and MCL achieved a performance increase of 1.1%, 1.0%, and 0.9% on VQA-RAD, PathVQA, and SLAKE dataset, respectively.

**Table 2.** Ablation Study on Different Pre-training Objective Settings.

| Training tasks | VQA-RAD (Overall) | PathVQA (Overall) | SLAKE (Overall) |
|---|---|---|---|
| ITM+MLM | 74.5 | 61.5 | 82.0 |
| ITM+MLM+UCL | 77.3 | 63.5 | 83.2 |
| ITM+MLM+MCL | 78.1 | 64.1 | 84.0 |
| **MUMC(ITM+MLM+UCL+MCL)** | **79.2** | **65.1** | **84.9** |

Furthermore, to assess the performance of the proposed masked image strategy and identify the optimal masking probability, experiments were conducted by varying the masking probabilities of input images at levels of 0%, 25%, 50% and 75%. As presented in Table 3, the results are consistent among all the three datasets. With 25% masking probability, the model yielded the best results, compared to no masking applied. The performance decreased if 50% and 75% masking probabilities were used.

**Table 3.** Study of different masked image probabilities.

| Masking probability | VQA-RAD (Overall) | PathVQA (Overall) | SLAKE (Overall) |
|---|---|---|---|
| 75% | 76.9 | 63.4 | 82.6 |
| 50% | 78.6 | 64.3 | 83.7 |
| **25%** | **79.2** | **65.1** | **84.9** |
| 0% | 77.8 | 64.0 | 83.2 |

### 3.5 Visualization

We utilized Grad-CAM [14] to visualize the cross-attention maps between the questions and images, and analyzed the relevance of the attended image regions for generating the answers. In Fig. 2, it showed some attention maps that overlayed on the original images. For answering open-ended questions, the model accurately attended to the relevant infarct regions, as shown in Fig. 2a and Fig. 2b. In Fig. 2a, to answer the question, "Where is/are the infarct located?", the model highlighted the areas that well covered the infarction. Interestingly, the model attended to infarct areas on both hemispheres (Fig. 2b) and generated the answer, "Bilateral". Besides the position-related questions, in Fig. 2c, it showed the attention map to answer the closed form question, "Is there any region in the brain that is lesioned?". The model successfully attended to the lesion area and provided the correct answer of "Yes". Moreover, the model demonstrated its ability to attend to the regions of ribs to answer the counting-related question in Fig. 2d, where the question was "Are there more than 12 ribs?", and the model accurately outputted the answer "Yes".

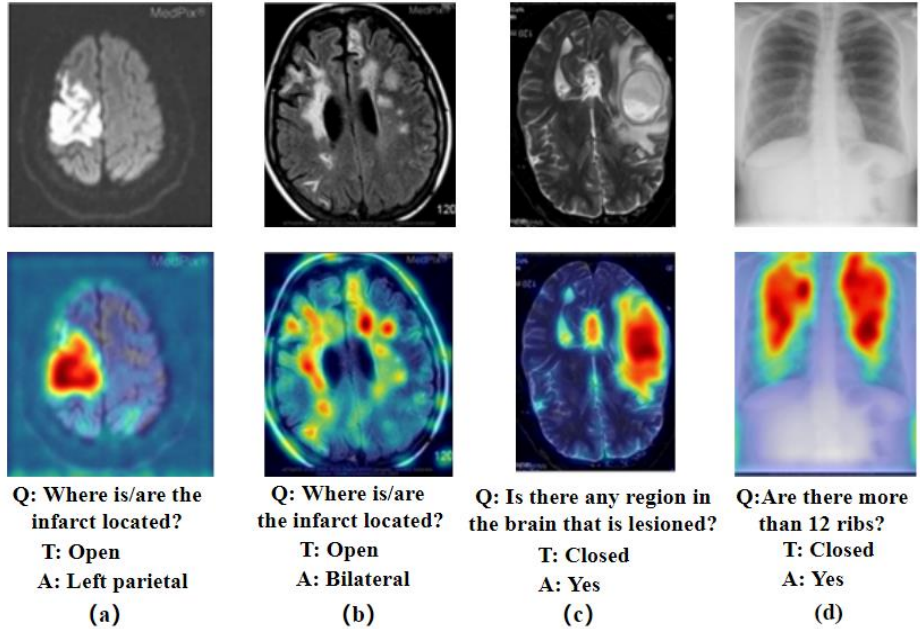

**Fig. 2.** Visualizations of the image attention maps on medical VQA tasks.

## 4      Conclusion

In this paper, we propose a new method to tackle the challenge of medical VQA tasks, which is pre-trained on the medical image caption datasets and then transferred to the downstream medical VQA tasks. The proposed self-supervised pre-training approach with unimodal and multimodal contrastive losses leads to significant performance improvement on three public VQA datasets. Also, using masked images as a data augmentation technique is proven to be effective for learning representations on medical visual and language tasks. As a result, our proposed method not only outperformed the state-of-the-art methods by a significant margin, but also demonstrated the potential for model interpretability.

**Acknowledgement.** This work is supported by Natural Science Foundation of Heilongjiang Province under grant number LH2021F015.